\documentclass[pdflatex,sn-mathphys-num]{sn-jnl}

\usepackage{graphicx}%
\usepackage{multirow}%
\usepackage{amsmath,amssymb,amsfonts}%
\usepackage{amsthm}%
\usepackage{mathrsfs}%
\usepackage[title]{appendix}%
\usepackage{xcolor}%
\usepackage{textcomp}%
\usepackage{manyfoot}%
\usepackage{booktabs}%
\usepackage{algorithm}%
\usepackage{algorithmicx}%
\usepackage{algpseudocode}%
\usepackage{listings}%
\usepackage{tikz}
\usetikzlibrary{shapes,arrows,positioning}
\usepackage{mathrsfs}
\usepackage{cleveref}
\usepackage{mathrsfs}
\UseRawInputEncoding
\theoremstyle{thmstyleone}%
\theoremstyle{thmstyletwo}%

\theoremstyle{thmstylethree}%

\begin{document}

\title[Attentive Reasoning Queries]{\centering Attentive Reasoning Queries: \\ A Systematic Method for Optimizing Instruction-Following in Large Language Models}

\author*{\fnm{Bar} \sur{Karov}}\email{bar@emcie.co}
\author{\fnm{Dor} \sur{Zohar}}\email{dor@emcie.co}
\author{\fnm{Yam} \sur{Marcovitz}}\email{yam@emcie.co}

\affil{\orgdiv{NLP Research}, \orgname{Emcie Co Ltd}}


\abstract{We present Attentive Reasoning Queries (ARQs), a novel structured reasoning approach that significantly improves instruction-following in Large Language Models through domain-specialized reasoning blueprints. While LLMs demonstrate remarkable capabilities across diverse tasks, they often fail to maintain adherence to complex, use-case-specific instructions during multi-turn conversations, presenting challenges for business-critical applications. ARQs address this limitation by guiding LLMs through systematic reasoning steps with targeted queries that reinstate critical instructions and facilitate intermediate reasoning throughout the completion process. In extensive testing within Parlant, our framework for reliable customer-facing agents in which ARQs were born out of necessity, they achieved a 90.2\% success rate across 87 test scenarios, outperforming both Chain-of-Thought reasoning (86.1\%) and direct response generation (81.5\%). ARQs showed particular strength in addressing persistent failure modes like guideline re-application and hallucination prevention. Our analysis also revealed that ARQs can potentially be more computationally efficient than free-form reasoning when carefully designed. These findings demonstrate that structured reasoning approaches provide effective mechanisms for controlling how LLMs process information and make decisions in complex scenarios.}

\keywords{Large Language Models, Attentive Reasoning Queries, Structured Reasoning, Conversational AI, Chain-of-Thought, Reasoning Framework, Reasoning Optimization, Hallucination Prevention, Customer-facing AI}



\maketitle 

\noindent \textbf{Supplementary Materials:} Source code, prompt examples and other supplementary materials are available on our~\href{https://github.com/emcie-co/parlant/tree/arqs-a-systematic-method-for-optimizing-instruction-following-in-llms}{GitHub}.

\section{Introduction}\label{sec1}

Large Language Models (LLMs) have demonstrated remarkable capabilities across diverse tasks, from knowledge retrieval to creative content generation~\cite{zhu2023large, chang2024survey}. However, ensuring that these models perform systematic, reliable reasoning—particularly in multi-turn conversational settings—remains challenging~\cite{zhang2025survey}. LLMs often struggle with hallucinations, remembering instructions, and maintaining consistent reasoning patterns across complex tasks. These challenges are especially pronounced in high-stakes customer-facing applications, such as a bank's customer service where a dynamic and temporal understanding of the context, and adherence to specific behavioral guidelines in relation to it, are critical.

Traditional approaches to enhancing LLM reasoning, such as free-form chain-of-thought prompting~\cite{wei2022chain} or step-by-step instruction, have shown promise but offer limited control over how models process information. While these methods encourage models to "think aloud," they provide minimal structure to guide the reasoning process through domain-specific considerations or known failure modes. Furthermore, as conversation context grows, LLMs often fail to maintain focus on critical instructions and constraints that should govern their behavior~\cite{dong2023bamboo, li2024long}.

In this paper, we introduce Attentive Reasoning Queries (ARQs), a structured approach to guide LLMs through systematic reasoning steps using targeted, task-specific queries. ARQs leverage domain knowledge to redirect the model's attention to critical instructions, decisions, and potential pitfalls at the points where such attention is most crucial. This approach serves two key functions: (1) Reinstating important instructions, (2) Facilitating intermediate reasoning steps. These functions are particularly instrumental in complex and nuanced conversational contexts in which adherence to specific instructions is essential.

We implement and evaluate ARQs within Parlant~\cite{parlant}, a framework for developing reliable conversational AI agents suitable for business-specific customer-facing applications. This framework requires agents to maintain strict adherence to behavioral guidelines, appropriately utilize available tools, and avoid hallucinations. By structuring the reasoning process through predefined JSON schemas with targeted queries, we test how ARQs can enhance performance across key processing components.

\section{Related Work}
\subsection{Prompting Techniques for Reasoning}
Many recent advances in LLM reasoning capabilities have been driven by specialized prompting techniques. Chain-of-Thought (CoT) prompting \cite{wei2022chain} demonstrated that eliciting intermediate reasoning steps before producing answers significantly improves performance on complex tasks. Chain-of-Verification (CoVe) \cite{dhuliawala2023chain} extends this approach by having models explicitly verify their outputs against potential errors. Other variations include zero-shot CoT \cite{kojima2022large} using simple prompts like ``Let's think step by step'' and Tree-of-Thought (ToT) approaches \cite{yao2023tree} that explore multiple reasoning pathways.

While effective, these general prompting strategies provide limited guidance on task-specific reasoning steps. They depend largely on the model's internal capabilities to determine appropriate reasoning patterns and typically lack domain-specific guidance that could prevent common failure modes or highlight critical considerations.

\subsection{Conversational Agents}
Conversational agents built on LLMs often incorporate additional structures to maintain coherence, adhere to guidelines, and extend functionality. ReAct \cite{yao2023react} pioneered the integration of reasoning and action, while frameworks like LangChain \cite{chase2022langchain} provide infrastructure for tool use and workflow management.

These frameworks often treat the core reasoning process as a black box, with limited mechanisms to guide how the LLM processes information or makes decisions. This creates challenges in ensuring consistent adherence to complex behavioral guidelines, particularly in business-critical customer-facing applications. Recent work highlights both the importance of explicit planning mechanisms and the challenges of tool integration~\cite{sypherd2024practical}. Studies show LLMs often struggle to select appropriate tools or provide correct parameters~\cite{shen2024llm}.

To address these challenges, recent advancements have focused on enhancing the interpretability and control of LLMs within conversational agents. For instance, the development of LangGraph~\cite{langgraph2025} extends LangChain's capabilities by introducing an orchestration framework for building stateful agents. Such approaches seek to address the difficulty of maintaining attention to numerous complex instructions by allowing more granular control over agent processing stages, restricting the set of provided instructions to the conversational contexts in which they are most relevant.

Parlant~\cite{parlant} is a new open-source framework which addresses these challenges through managed, supervised guidelines. It implements a dynamic control system that evaluates conversational context against a structured repository of behavioral guidelines. The framework follows a modular architecture with specialized processing components for guideline filtering and matching, tool calling, and message generation, each operating with explicit reasoning protocols using ARQs. In this work, we use Parlant's test suite as a controlled environment to evaluate whether ARQs can improve reasoning performance over CoT in complex and nuanced conversational contexts.

\subsection{Common Pitfalls in LLM-based Systems}
LLMs exhibit several persistent failure modes in conversational applications that limit their reliability. Some notable pitfalls are:
\begin{itemize}
\item \textbf{Alignment drift} occurs when models gradually deviate from specified guidelines over extended conversations~\cite{shen2023large}.

\item \textbf{Hallucination}— the generation of factually incorrect information, or the offering of unwarranted services, represents a related challenge to alignment drift~\cite{huang2025survey}. As models drift from alignment constraints, they tend to generate content lacking proper grounding in the provided context. This is particularly problematic for domain-specific agents that must maintain factual accuracy within their expertise areas.
\item \textbf{Context forgetfulness} manifests as pronounced recency bias \cite{liu2024lost}, where models preferentially attend to information near the end of their context window while neglecting earlier content. This creates significant challenges in extended conversations where important instructions or context may be buried among various exchanges and constraints.
\end{itemize}
\section{Our Contributions}
This paper makes the following contributions:
\begin{enumerate}
    \item \textbf{ARQ Methodology:} We introduce Attentive Reasoning Queries (ARQs) as a structured approach to guide LLM reasoning, which demonstrate how domain knowledge can be incorporated into reasoning blueprints to address task-specific challenges and known failure modes.
    \item \textbf{ARQ Implementation:} We implement ARQs within Parlant, a framework for developing conversational agents supporting strict quality requirements with respect to agent behavioral patterns and tool-use. 
    \item \textbf{Empirical Evaluation:} We conduct an evaluation comparing ARQ performance against both Chain-of-Thought and no-reasoning (control) approaches, demonstrating that ARQs achieve superior performance, particularly in addressing challenging failure modes such as guideline re-application and hallucination prevention.
\end{enumerate}

The dataset and code used to produce our experiment is available on our GitHub (see title page), along with the full code for the Parlant framework.  

\section{Attentive Reasoning Queries (ARQ)}
As a motivating example, consider how we might choose a restaurant for a group dinner. Our decision process follows a structure. First, we review the available restaurants in terms of dietary preferences, budget, and location. Next, we evaluate each individual's preferences with respect to these options.

Importantly, this pattern of systematic review followed by structured evaluation can be abstracted and applied across many similar tasks.

This reasoning process inspires our development of Attentive Reasoning Queries (ARQs). ARQs guide Large Language Models (LLMs) through systematic reasoning blueprints. This is achieved by requiring responses to follow a predefined JSON schema in which keys are pre-defined, pinpointed queries designed to direct the model's attention to relevant information. The values are then filled by an LLM during its response completion process.

These queries can either be general-purpose or domain-specific, depending on the task. For example, for the described task, a reasonable reasoning chain would be answering the following questions before coming to a decision:
\begin{itemize}
    \item What are the constraints of the group (dietary, financial, or otherwise)?
    \item What restaurants are open and within range?
    \item For each of these restaurants, how suitable is their menu to our constraints?
\end{itemize}

\begin{figure}[!t]
\centering
\includegraphics[width=1.0\textwidth]{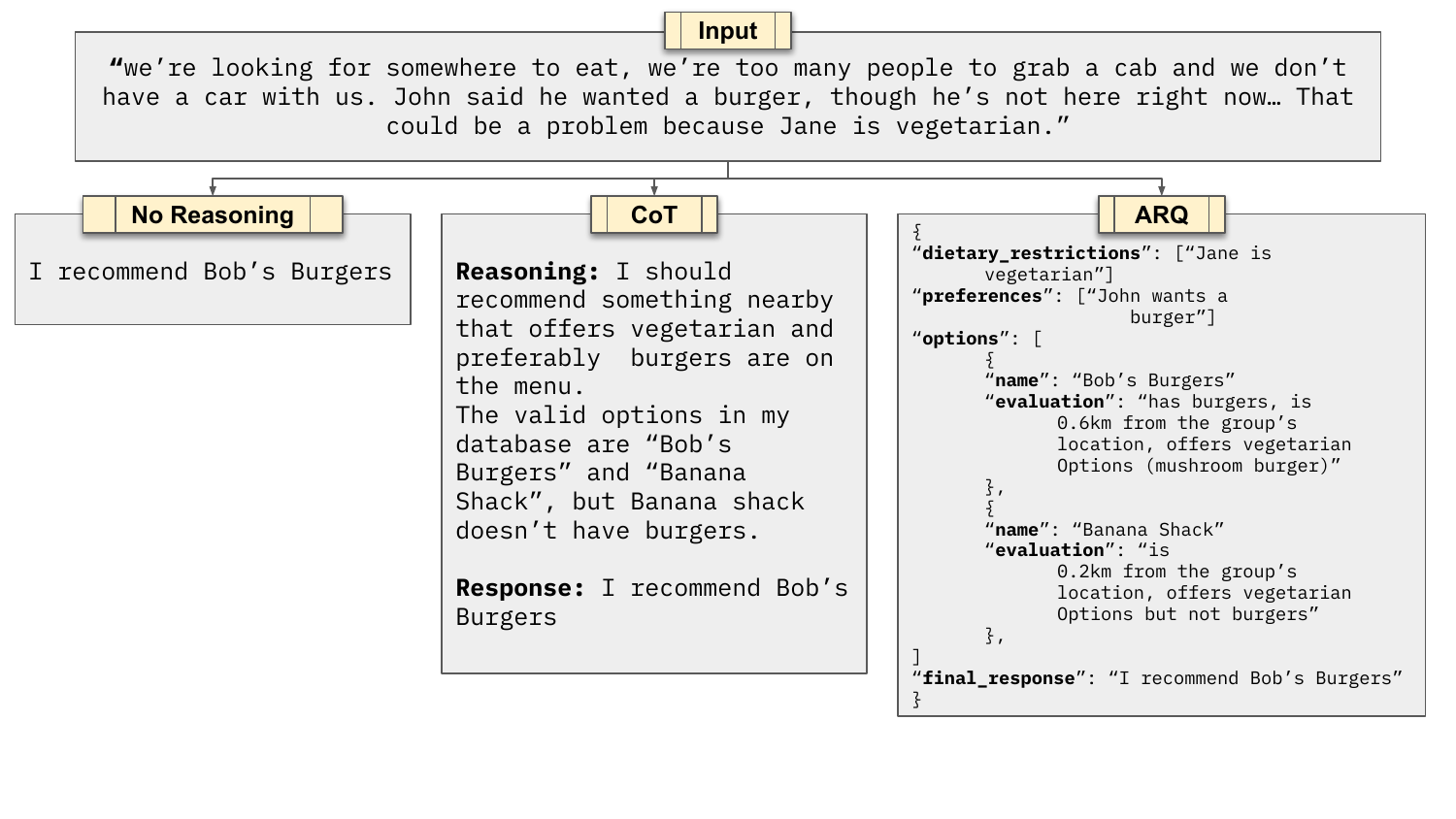}
\caption{Output examples: (1) using ARQs, (2) using CoT, and (3) performed without reasoning, all leading to the same final output. We assume the agent is equipped with information about nearby restaurants and their menus.}
\label{fig:arq-example}
\end{figure}

Given this desired reasoning blueprint, we can prompt an LLM to respond to these queries before arriving at its final recommendation. This is accomplished by instructing the model to return a JSON object (or another structured format compatible with the chosen LLM) containing evaluations for each query in the sequence. See~\Cref{fig:arq-example} for a simple implementation example.

This structured approach also makes extracting the model's final answer easier, compared to other reasoning methods, as conclusions appear in specific query responses rather than within lengthy reasoning text, simplifying both human review and automated processing.

The ARQ-guided process consists of the following steps:
\begin{enumerate}
\item \textbf{Leading ARQ Phase:} The LLM processes a sequence of pre-determined leading questions that serve three key functions:
\begin{itemize}
\item Reinstating critical instructions
\item Reinstating important contextual information from the prompt
\item Facilitating step-by-step reasoning and intermediate computations
\end{itemize}
\item \textbf{Response Generation:} Based on the reasoning from the leading query phase, the LLM produces a response.
\item \textbf{(Optional) Response Verification:} The LLM evaluates if its suggested response satisfies all requirements, by answering pre-defined verification questions (E.g, "Is the response consistent with the constraints listed in the leading ARQ stage?"). If not, a new response candidate is generated and verified, until a satisfactory one is found.
\end{enumerate}

\subsection{Query Design}
By leveraging domain knowledge about specific tasks, ARQs can be designed to target known failure modes and critical reasoning steps. Unlike free-form CoT or CoVe approaches that rely on generic prompting strategies for nonspecific use cases, ARQs can be crafted to address task-specific challenges and guide the LLM through sensitive decision points. 

For example, in information retrieval tasks, ARQs can guide the LLM to systematically identify relevant sources, assess their reliability, cross-reference claims across multiple documents, and check for temporal consistency and contextual relevance before synthesizing a response. 

Critical ARQs can be identified either through analytic decomposition of a complex reasoning process, or, better yet, through experimentation and feedback. An additional important benefit of ARQs emerges in practice, as their corresponding completions often help identify potential contradictions, misinterpretations, or lack of contextual grounding that might be overlooked in open-ended reasoning methods. 

ARQs leverage a key characteristic of LLM behavior—the enhanced recall of information that appears near the end of the input context. ARQs are designed to make the LLM reiterate critical instructions using the leading-query completions just-in-time, right before attending to the completion of the required output. Leveraging these properties of autoregressive models, this recency effect, wherein the LLM reiterates critical information at the end of its context window, empirically helps maintain important constraints and requirements in the LLM's active context during response generation.

\begin{figure}[ht]
\centering
\includegraphics[width=0.7\textwidth]{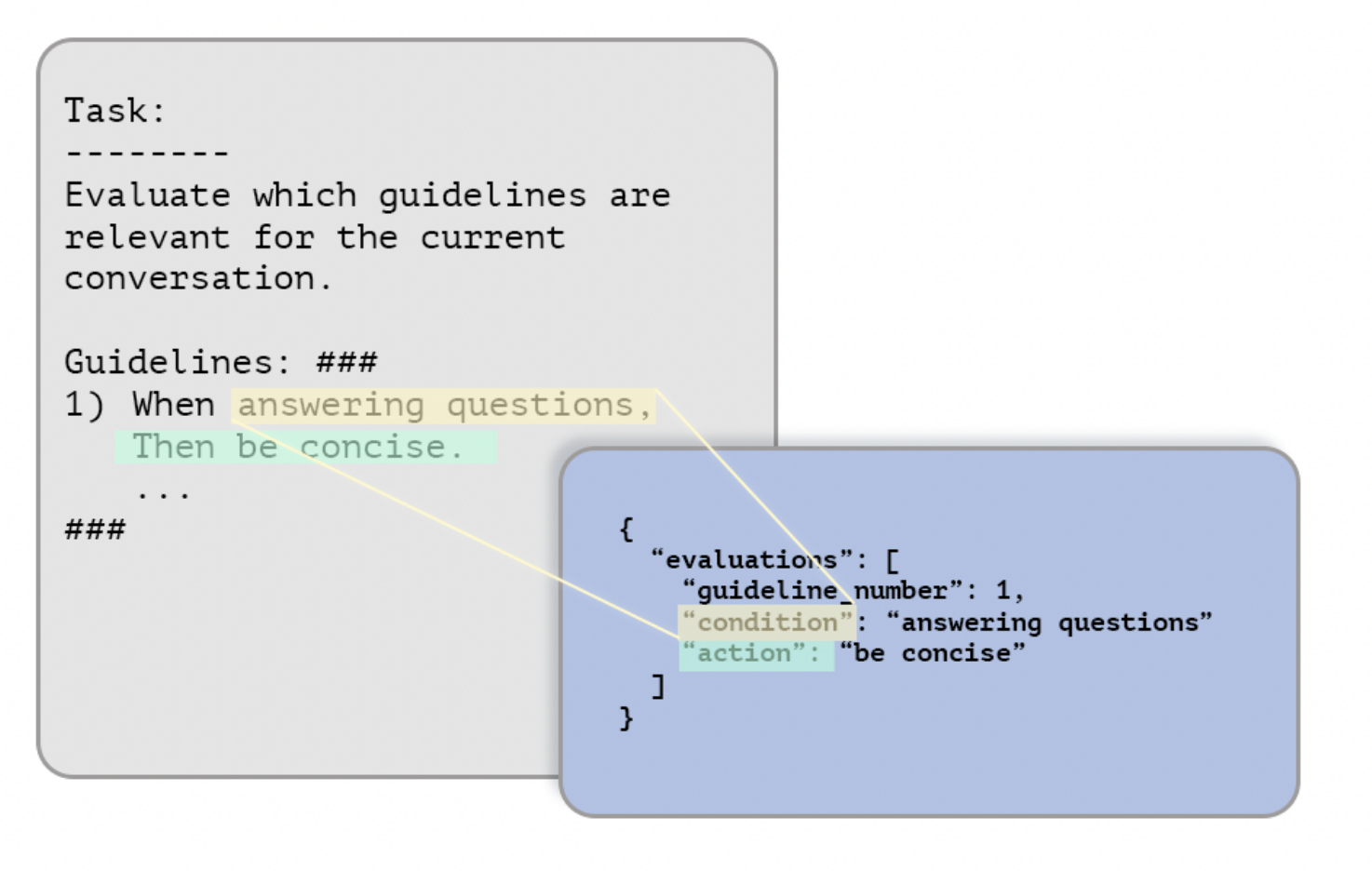}
\caption{An illustration of query-attention hypothesis: an ARQ asking the LLM to parrot the guideline before evaluating its relevance}
\label{fig:parroting-arq}
\end{figure}

This approach may also provide additional benefits through the LLM's attention mechanism. Specifically, we hypothesize that responding to leading queries which ask the LLM to repeat critical instructions allows the LLM to highlight and thus establish stronger attention patterns between task-specific input (such as different instances of user queries) and general instructions (such as how to handle user queries).  See~\Cref{fig:parroting-arq} for an example of such an ARQ. However, a detailed investigation of this attention-based hypothesis falls outside the scope of this paper.

\section{Setting} \label{sec:setting}
To evaluate the efficacy of Attentive Reasoning Queries (ARQs) in real-world applications, we test them in the scope of a Conversational AI engine for customer-service use cases—a domain where systematic reasoning and adherence to specific guidelines can be particularly challenging and consequential.

The reference engine design described here forms part of Parlant, a framework for developing reliable conversational AI agents suitable for customer-facing applications. 

In Parlant, each agent is initialized with four key components provided by its designer:
\begin{enumerate}
    \item \textbf{Agent Profile:} A concise, natural-language description defining the agent's purpose and operational scope.
    
    \item \textbf{Behavioral Guidelines:} A collection of conditional instructions in the form of ``When $\langle$X$\rangle$ Then $\langle$Y$\rangle$'' statements (e.g., ``When discussing weather Then use metric units'').
    
    \item \textbf{Tool Suite:} A set of external functions accessible via structured API methods, enabling the agent to retrieve information or execute actions within its environment.
    
    \item \textbf{Domain Lexicon:} A comprehensive glossary of domain-specific terms, essential to the agent's operational context.
\end{enumerate}

The agent engages in multi-turn conversations while maintaining four critical constraints: (1) Strict adherence to its prescribed guidelines, (2) Appropriate utilization of available tools, (3) Accurate application of information provided in its profile and domain lexicon, and (4) Elimination of hallucinated information or services not explicitly authorized by its designer.

 \subsection{Engine Architecture}
\begin{figure}[h]
\centering
\includegraphics[width=1.0\textwidth]{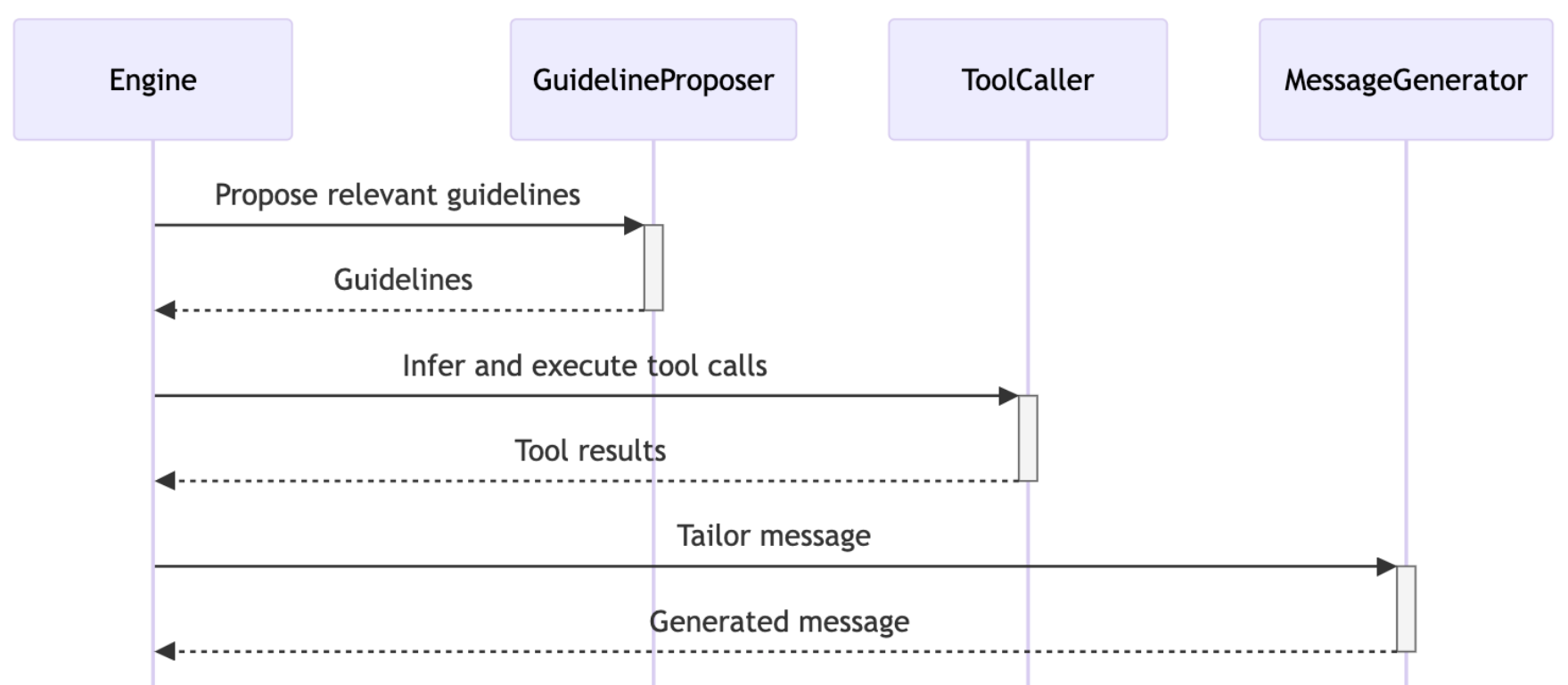}
\caption{A diagram of the Parlant Engine, including how its different modules interact. Modules execute from left to right, where each module feeds the next one with its outputs. The message generator receives comprehensive inputs from all other modules, before tailoring a final response.}
\label{fig:parlant-engine}
\end{figure}
To fulfill these requirements, agent responses undergo a modular processing pipeline with specialized LLM calls using predefined prompt templates. When processing a user message, the agent executes the following sequence, as shown in~\Cref{fig:parlant-engine}:
\begin{itemize}
    \item \textbf{Guideline Proposition:} Identifies which guidelines are applicable to the current state of the conversation.
    
    \item \textbf{Tool Calling:} Determines optimal tool selection and parameter configuration based on user intent and conversation state.
    
    \item \textbf{Message Generation:} Crafts a final response incorporating selected guidelines and tool outputs.
\end{itemize}

Guideline proposition and tool calling operate in an iterative cycle, as tool call results may activate additional guidelines. For example, if a geolocation tool identifies that a user is located in Europe, this might trigger a guideline of the form ``When the user is from Europe Then use metric units.''

Additionally, each tool in the system is explicitly attached to at least one guideline, creating a controlled structure that governs tool access. A tool can only be called if its associated guideline has been determined to be active in the current context. This constraint ensures that tool usage aligns with the designer's intent and prevents inappropriate actions.

\subsubsection{Guideline Proposition} \label{sec:guideline_proposer}
A guideline in our framework consists of a condition (the ``When'' clause) and an action (the ``Then'' clause). The Guideline Proposer module determines which guidelines should be active given the current conversation state. This determination is more nuanced than simple condition matching and requires contextual reasoning.

The Guideline Proposer receives as input:
\begin{enumerate}
    \item \textbf{Interaction History:} All past messages and tool call results.
    \item \textbf{Agent Profile \& Domain Lexicon:} The agent's role description and domain-specific terminology.
    \item \textbf{Staged Tool Calls:} Results from recently executed tools.
    \item \textbf{Guidelines to Filter:} The complete set of available guidelines.
\end{enumerate}
It then outputs a score from 1 to 10, indicating how strongly each guideline currently applies. Guidelines with a score of 6 or more are activated. This score is also carried over to the Message Generator, using it to prioritize possibly conflicting guidelines.

Guideline activation follows a decision process that considers temporal context. To see why simple condition-matching is not sufficient, consider two guidelines with identical conditions: ``When the customer is ordering a pizza Then offer a 2-for-1 special'' and ``When the customer is ordering a pizza Then never recommend pineapple topping.'' The first should activate once and then become inactive after execution, while the second must remain active throughout the ordering process.

To handle these nuances, we implement the following activation protocol:
\begin{enumerate}
    \item \textbf{Condition Evaluation:} Determine if the guideline's condition applies to the current context.
    \item \textbf{Continuity Assessment:} Classify the action as either one-time or continuous. Continuous actions (like maintaining a specific tone or avoiding particular recommendations) remain active as long as their condition applies.
    \item \textbf{Previous Application Check:} For non-continuous actions, verify if the action has already been performed.
    \item \textbf{Reactivation Analysis:} For non-continuous actions previously performed, determine if the condition became false and then true again, warranting reactivation.
\end{enumerate}

Our implementation includes a few additional nuanced rules for guideline re-application that address edge cases such as partially fulfilled multi-part actions and assessing action parts as either cosmetic or functional in terms of their impact on the conversation. For brevity, these details are included in~\Cref{app:guideline-proposer}, along with the guideline propoesr ARQ design used in Parlant.

\subsubsection{Tool Calling}
The Tool Caller module is responsible for determining which tools should be executed given the currently active guidelines. 

The Tool Caller receives the following inputs:
\begin{enumerate}
    \item \textbf{Interaction History:} All past messages and tool call results.
    \item \textbf{Agent Profile \& Domain Lexicon:} Contextual information about the agent's role and terminology.
    \item \textbf{Active Guidelines:} The set of guidelines determined to be active by the Guideline Proposer.
    \item \textbf{Available Tools:} Tools attached to active guidelines, each with its own parameter requirements and descriptions.
    \item \textbf{Staged Tool Calls:} Results from tools already executed in the current processing cycle.
\end{enumerate}

The Tool Caller follows specific instructions, provided to it in its prompt, to determine when and how tools should be activated, focusing on both contextual relevance and practical considerations:

\begin{enumerate}
    \item \textbf{Proactive Tool Usage:} Tools may be suggested even when they don't directly address the user's latest message, if they could advance the conversation to a more productive state.
    
    \item \textbf{Multiple Invocations:} Each tool may be called multiple times with different arguments within a single response cycle. For example, a product search tool might be called separately for each product category mentioned by the user.
    
    \item \textbf{Call Duplication Prevention:} The system avoids calling a tool with identical arguments more than once, unless there is a clear justification, such as refreshing potentially outdated information.
    
    \item \textbf{Dependency Management:} Each tool call is designed to be self-contained, relying only on information available in the immediate context or from already-executed tool calls. This prevents dependency chains that could fail when executed in parallel.
\end{enumerate}

To further understand the tool caller's process, and to examine the ARQs it uses in Parlant, see~\Cref{app:tool-caller}.

\subsubsection{Message Generation} \label{sec:message-generator}
The Message Generator is the final module in the processing pipeline, responsible for synthesizing the outputs from previous stages into a coherent, contextually appropriate response to the user.

The Message Generator receives the following inputs:
\begin{enumerate}
    \item \textbf{Interaction History}
    \item \textbf{Agent Profile \& Domain Lexicon} \item \textbf{Active Guidelines:} Active guidelines, as determined by the guideline proposer module.
    \item \textbf{Tool Call Results:} Data retrieved from any tools executed by the tool caller.
\end{enumerate}

The Message Generator is provided with several preset instructions in its prompt that apply to all interactions, independent of the specific guidelines it receives. These instructions ensure consistent, high-quality responses across varied conversation contexts and cover aspects such as communication best-practices, information handling, and presentation format. For the complete set of instructions provided to the Message Generator and its ARQ implemenation, see either~\Cref{app:message-generator}, or the full Message Generator prompt in the supplementary materials on Github.

The Message Generator also handles guideline prioritization during response synthesis rather than in the earlier guideline proposition stage. This design choice avoids additional cross-batch LLM calls that would increase system latency. When resolving conflicts, the module considers both the applicability scores assigned during guideline proposition and specific conflict resolution instructions provided in its prompt.

To prevent hallucination, the Message Generator is explicitly instructed to offer only services and information explicitly provided in its context. This constraint ensures the agent doesn't suggest unauthorized options—for instance, a pizza delivery agent won't propose self-pickup unless this service was specifically included by the designer.

\section{Experiment}\label{sec:experiment}

\subsection{Experimental Design}\label{sec:exp-design}

To empirically evaluate the effectiveness of ARQs, we implemented them within the Parlant framework described in Section~\ref{sec:setting}. Our experiments were designed to compare ARQ performance against alternative reasoning approaches when deployed across the three core modules of our system: Guideline Proposer, Tool Caller, and Message Generator.

For each module, we developed three methodologically distinct implementations:

\begin{enumerate}
    \item \textbf{ARQ Implementation}: Employs the structured query-based reasoning approach described in Section~\ref{sec1}, with module-specific queries designed to target known failure modes and critical decision points in each component. Care was taken to ensure that ARQs are general rather than overfitting the test cases.
    
    \item \textbf{Chain-of-Thought (CoT) Implementation}: Incorporates free-form reasoning before generating the final output, allowing the LLM to develop its own reasoning pathway without the structured constraints of ARQs.
    
    \item \textbf{Control Implementation}: Generates direct responses based on instructions, without any explicit reasoning process.
\end{enumerate}

All implementations share identical base prompts, receiving the same instructions and functional requirements.

The evaluation framework is publicly available, including the full dataset of test scenarios, detailed evaluation criteria, and implementation code for all three reasoning methodologies. For further details see the attached Github repository.

\subsubsection{In-Context Learning}\label{subsubsec:icl}

To optimize performance across all implementations, we incorporated In-Context Learning (ICL) through carefully selected exemplars. Each module's prompt includes a set of few-shot examples that demonstrate successful execution patterns. These examples were iteratively refined based on observed failure modes in real customer interactions, and are identical across all 3 reasoning methods.

For the ARQ implementation, the examples include both the structured queries and their corresponding responses, modeling the expected reasoning pattern. For CoT and Control implementations, only the expected response is provided.

\subsection{Evaluation Dataset}\label{subsec:dataset}

Our evaluation utilized a comprehensive dataset of 87 test cases, crafted to assess the system's adherence to framework requirements under diverse conversational scenarios. The dataset composition includes 22 scenarios focused exclusively on guideline proposition accuracy, and additional 65 comprehensive scenarios evaluating the full interaction pipeline (guideline proposition, tool calling, and response generation)

Each test case provides:
\begin{enumerate}
    \item \textbf{Agent Configuration}: Profile description, behavioral guidelines, available tools, and domain lexicon
    \item \textbf{Conversation History}: A sequence of user/agent interactions leading to the current state, ending with a user message to respond to
    \item \textbf{Success Criteria}: Conditions that must be satisfied for the response to be considered correct and aligned
\end{enumerate}

Response quality was assessed through multiple complementary approaches. Our primary evaluation used an LLM to judge whether responses met the predefined success criteria for each test case, examining both content accuracy (e.g., an LLM determines whether the agent's response satisfies the criteria ``includes an offering of a 10\% discount''). For scenarios requiring external tool usage, we evaluated whether the agent correctly identified when tools were needed, selected the appropriate tools from its available set, and provided the necessary parameters for successful execution.

Tests that apply exclusively to the guideline proposer were evaluated based on the set of guidelines that the guideline proposer suggested. A guideline proposer test is considered successful only if the correct set of guidelines is proposed.

\subsection{Language Model}
All experiments were performed using OpenAI's GPT-4o model family~\cite{openai2024gpt4o}, with specific versions selected based on extensive testing in real-world Parlant applications. We used the gpt-4o-2024-11-20 version for the Tool Caller module and the gpt-4o-2024-08-06 version for both the Guideline Proposer and Message Generator modules. Temperature settings were configured as follows: 0.1 for the Message Generator, 0.15 for the Guideline Proposer, and 0.05 for the Tool Caller.

\subsection{Results}\label{subsec:results}
We conducted experiments comparing the performance of each reasoning method (Control, Chain-of-Thought, and ARQ) across all three modules in our framework. Each test in the dataset was run 5 times to account for the stochastic nature of LLM outputs. 

\begin{figure}[ht]
\centering
\begin{tabular}{l|c|c|c}
\toprule
\textbf{\begin{tabular}[c]{@{}l@{}}Reasoning\\Method\end{tabular}} & 
\textbf{\begin{tabular}[c]{@{}c@{}}Guideline\\Proposer Tests (\%)\end{tabular}} & 
\textbf{\begin{tabular}[c]{@{}c@{}}Comprehensive\\Tests (\%)\end{tabular}} & 
\textbf{\begin{tabular}[c]{@{}c@{}}Total\\(\%)\end{tabular}} \\
\midrule
None & 70.43 & 85.31 & 81.54 \\
CoT     & 80.87 & 87.81 & 86.05 \\
ARQ     &  \textbf{84.24} & \textbf{92.19} & \textbf{90.17} \\
\bottomrule
\end{tabular}
\caption{Performance comparison across reasoning methods. Comprehensive tests evaluate all three modules (message generator, tool caller, and guideline proposer) working in conjunction.}
\label{fig:performance-comparison}
\end{figure}

As shown in Figure~\ref{fig:performance-comparison}, ARQs achieved the highest success rate on our dataset, outperforming both Chain-of-Thought and the Control setting where no reasoning was performed during the completion stage. 

Our analysis revealed that tests passed exclusively by ARQ (failing under CoT) generally fall into two categories:
\begin{itemize}
    \item \textbf{Guideline re-application:} Tests requiring nuanced decisions about the re-activation of guidelines that were previously followed in the agent's earlier responses. 
    
    \item \textbf{Hallucination prevention:} Tests specifically designed to detect whether the agent offers hallucinated facts or services not supported by its available tools or context.
\end{itemize}

Based on our experience deploying conversational agents in production environments, these two failure cases represent some of the most challenging adherence issues for LLM-based systems. This fact highlights the ability of ARQs to target critical fail-points in the decision process, as structured reasoning queries can be strategically designed and added to address the most persistent weaknesses in the decision process.

\subsubsection{Computational Efficiency}\label{subsubsec:computation}

\begin{figure}[ht]
\centering
\begin{tabular}{l|c|c|c}
\toprule
\textbf{Module} & \textbf{Control} & \textbf{CoT} & \textbf{ARQ} \\
\midrule
Message Generator & 54 & 330 & 596 \\
Tool Caller & 68 & 180 & 550 \\
Guideline Proposer & 48 & 405 & 289 \\
\bottomrule
\end{tabular}
\caption{Average output token usage by module and reasoning method.}
\label{fig:token-usage}
\end{figure}

Our analysis reveals that the specific design of ARQs and the nature of the underlying task significantly impact their computational efficiency relative to other reasoning methods, as measured by the output tokens required. As shown in Figure \ref{fig:token-usage}, token usage patterns vary across different modules, demonstrating that ARQs can be either more or less efficient than Chain-of-Thought depending on implementation choices and task characteristics.

The Guideline Proposer module demonstrates lower token usage with ARQs than with CoT, requiring 29\% fewer tokens while delivering superior performance. This efficiency stems primarily from the nature of the task- determining whether guidelines are active or inactive, which naturally lends itself to structured queries with concise responses. The task also has fewer edge cases compared to other modules, and does not require generating extensive natural language outputs.

In contrast, the ARQ implementations for the Message Generator and Tool Caller modules consumed substantially more tokens. These modules face more complex tasks requiring autoregressive natural language generation and handling numerous edge cases, which results in more extensive reasoning through ARQs.

This variation across modules underscores a critical finding: The efficiency of structured reasoning approaches depends on both how the queries are formulated and the inherent complexity of the underlying task. When queries direct the model to focus precisely on the most relevant aspects of a decision process within naturally bounded tasks like classification, they can reduce computational overhead. These findings suggest that ARQ design should be approached strategically based on the specific reasoning requirements and characteristics of each task.

\section{Limitations and Future Research}
While our experiments demonstrate the potential efficacy of ARQs in enhancing the reasoning capabilities of conversational AI agents, several limitations of the current study suggest important directions for future research.

Our evaluation focused specifically on conversational agents operating within the Parlant framework, leaving open questions about ARQ applicability in other contexts. Additionally, our current evaluation dataset, though carefully designed to test specific capabilities, remains modest in size and scope. Future work should validate ARQ performance on substantially larger and more diverse datasets of conversational scenarios that were not used during ARQ development.

The generalizability of our findings is constrained by our exclusive use of GPT-4o as the underlying language model. Preliminary research done in Parlant, which is not presented in this work, suggests that the results are reproducible across different models, though full empirical testing of this hypothesis remains a subject for future research.

Perhaps most significantly, our investigation primarily focused on establishing whether ARQs confer performance benefits in a conversational agent framework, without exploring the broader design space of ARQ construction or optimization strategies. We plan to conduct a systematic exploration of ARQ design principles in future work. This future research will establish formalized methodologies for constructing ARQs optimized for particular reasoning tasks or domains.

\section{Conclusion}
In this work, we introduced Attentive Reasoning Queries (ARQs), a structured approach to guide the reasoning processes of Large Language Models. ARQs utilize targeted, domain-specific questions organized within a predefined JSON schema to direct model attention to critical instructions and decision points. We implemented and evaluated ARQs within the Parlant framework, testing their effectiveness in conversational agent applications that require strict adherence to behavioral guidelines.
Our evaluation compared ARQs against Chain-of-Thought (CoT) reasoning, demonstrating that ARQs improves performance across the system's core modules.

\subsection{ARQs vs. Chain-of-Thought}
While both Chain-of-Thought and ARQs aim to enhance LLM reasoning capabilities, they differ fundamentally in their structure and implementation. CoT prompting encourages models to generate intermediate reasoning steps in a free-form manner before producing a final answer. This approach relies on the model's inherent capabilities with minimal external guidance.
In contrast, ARQs provide explicit structural scaffolding through predefined queries that guide the model's attention to specific objects during the reasoning process. This approach offers several advantages:

\begin{itemize}
    \item \textbf{Domain-Specific Guidance:} Unlike the general-purpose nature of CoT, ARQs incorporate domain knowledge to address task-specific challenges and known failure modes.
    \item \textbf{Enhanced Debuggability:} The structured format of ARQs allows system designers to more easily inspect and debug reasoning processes. When errors occur, designers can identify exactly which query or reasoning step fell short of the goal.
    \item \textbf{Attention Preservation:} ARQs strategically reinstate critical instructions and constraints at key decision points, addressing the "lost in the middle" phenomenon where important information receives less attention from the model.
\end{itemize}

\begin{figure}[!t]
\centering
\includegraphics[width=0.7\textwidth]{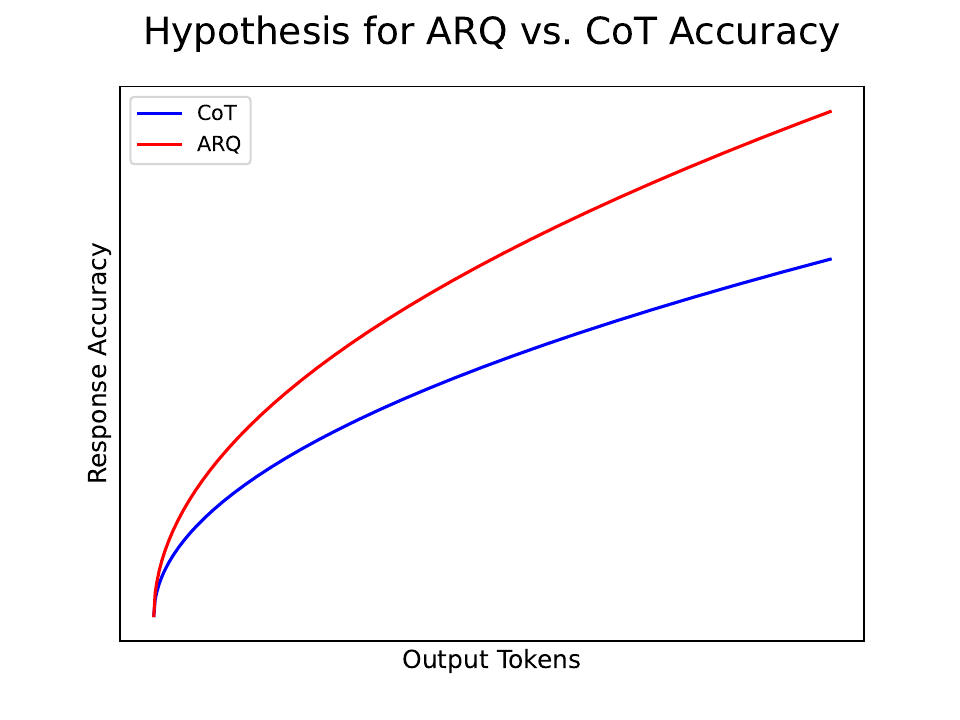}
\caption{Our hypothesis for how ARQ scales compared to CoT, as a factor of the reasoning length}
\label{fig:arq-cot-hypothesis}
\end{figure}

As shown in~\Cref{fig:arq-cot-hypothesis}, and in accordance with our practical experience, we hypothesize—though do not test within this paper—that as reasoning complexity increases, requiring more output tokens, both methods would show improved performance, but ARQs would likely scale more effectively. By explicitly directing the model's focus throughout extended reasoning chains, ARQs potentially avoid the degradation in reasoning quality that often occurs with longer free-form reasoning.

\bibliography{sn-bibliography}

\appendix
\section{Guideline Proposer Prompt and ARQs}
\label{app:guideline-proposer}
Examples for the full guideline proposer prompt, using all 3 reasoning modes, are available on our~\href{https://github.com/emcie-co/parlant/tree/arqs-a-systematic-method-for-optimizing-instruction-following-in-llms}{GitHub}.

The instructions regarding guideline application and re-application, as specified in the guideline proposer's prompt, are:
\begin{lstlisting}
GENERAL INSTRUCTIONS
-----------------
In our system, the behavior of a conversational AI agent is guided by "guidelines". The agent makes use of these guidelines whenever it interacts with a user (also referred to as the customer).
Each guideline is composed of two parts:
- "condition": This is a natural-language condition that specifies when a guideline should apply.
          We look at each conversation at any particular state, and we test against this
          condition to understand if we should have this guideline participate in generating
          the next reply to the user.
- "action": This is a natural-language instruction that should be followed by the agent
          whenever the "condition" part of the guideline applies to the conversation in its particular state.
          Any instruction described here applies only to the agent, and not to the user.


Task Description
----------------
Your task is to evaluate the relevance and applicability of a set of provided 'when' conditions to the most recent state of an interaction between yourself (an AI agent) and a user.
These conditions, along with the interaction details, will be provided later in this message.
For each condition that is met, determine whether its corresponding action should be taken by the agent or if it has already been addressed previously.


Process Description
-------------------
a. Examine Interaction Events: Review the provided interaction events to discern the most recent state of the interaction between the user and the agent.
b. Evaluate Condition/s: Assess the entire interaction to determine whether each condition is still relevant and directly fulfilled based on the most recent interaction state.
c. Check for Prior Action: Determine whether the condition has already been addressed, i.e., whether it applied in an earlier state and its corresponding action has already been performed.
d. Guideline Application: A guideline should be applied only if:
    (1) Its condition is currently met and its action has not been performed yet, or
    (2) The interaction warrants re-application of its action (e.g., when a recurring condition becomes true again after previously being fulfilled).

For each provided guideline, return:
    (1) Whether its condition is fulfilled.
    (2) Whether its action needs to be applied at this time. See the following section for more details.

Insights Regarding Guideline re-activation
-------------------
A condition typically no longer applies if its corresponding action has already been executed.
However, there are exceptions where re-application is warranted, such as when the condition is re-applied again. For example, a guideline with the condition "the customer is asking a question" should be applied again whenever the customer asks a question.
Additionally, actions that involve continuous behavior (e.g., "do not ask the user for their age", or guidelines involving the language the agent should use) should be re-applied whenever their condition is met, even if their action was already taken.
If a guideline's condition has multiple requirements, consider it continuous if at least one of them is continuous. Actions like "tell the customer they are pretty and help them with their order" should be considered continuous, since 'helping them with their order' is continuous.
Actions that forbid certain behaviors are generally considered continuous, as they must be upheld across multiple messages to ensure consistent adherence.

IMPORTANT: guidelines that only require you to say a specific thing are generally not continuous. Once you said the required thing - the guideline is fulfilled.

Conversely, actions dictating one-time behavior (e.g., "send the user our address") should be re-applied more conservatively.
Only re-apply these if the condition ceased to be true earlier in the conversation before being fulfilled again in the current context.

IMPORTANT: Some guidelines include multiple actions. If only a portion of those actions were fulfilled earlier in the conversation, treat the guideline as though it has been fully executed.
In such cases, re-apply the guideline only if its condition becomes true again later in the conversation, unless it is continuous.
\end{lstlisting}

When using ARQs, the guideline proposer is instructed to return a dictionary whose keys are pre-defined questions, and values are the LLM's responses to these questions. 

As an example, when evaluating the guideline:
\begin{itemize}
    \item \textbf{Condition:} a client asks for a drink
    \item \textbf{Action:} check if the drink is available in stock
\end{itemize}
We end the guideline proposer prompt by telling the LLM to return the following\footnote{See ending of 'ARQ Guideline Proposer Example Prompt.txt' in the supplementary materials to view in .txt format, for better readability}:
\begin{lstlisting}
{
    "guideline_id":"...",
    "condition":"a client asks for a drink",
    "condition_application_rationale":"<Explanation for why the condition is or isn't met>",
    "condition_applies":"<BOOL>",
    "action":"check if the drink is available in stock",
    "guideline_is_continuous":"<BOOL: Optional, only necessary if guideline_previously_applied is true. Specifies whether the action is taken one-time, or is continuous>",    "capitalize_exact_words_from_action
    _in_the_explanations_to_avoid_semantic_pitfalls":true,
    "guideline_previously_applied_rationale":{
        "<action_segment_1>":"<explanation of whether this action segment was already applied; to avoid pitfalls, try to use the exact same words here as the action segment to determine this. use CAPITALS to highlight the same words in the segment as in your explanation>",
        "<action_segment_N>":"<explanation...>"
    },
    "guideline_current_application_refers_to_a_new_or_subtly
    _different_context_or_information":"<if the guideline DID previously apply, explain here whether or not it needs to re-apply due to it being applicable to new context or information>",
    "guideline_previously_applied":"<str: either 'no', 'partially' or 'fully' depending on whether and to what degree the action was previously preformed>",
    "is_missing_part_cosmetic_or_functional":"<str: only included if guideline_previously_applied is 'partially'. Value is either 'cosmetic' or 'functional' depending on the nature of the missing segment.",
    "guideline_should_reapply":"<BOOL: Optional, only necessary if guideline_previously_applied is not 'no'>",
    "applies_score":"<Relevance score of the guideline between 1 and 10. A higher score indicates that the guideline should be active>"

}
\end{lstlisting}
Where text in angled brackets represents our instruction to the LLM, rather than actual text it has to output.

These ARQs guide the LLM through an assessment of whether:
\begin{enumerate}
\item The condition currently applies to the conversation state
\item The action has been previously performed (fully, partially, or not at all)
\item The guideline represents continuous or one-time behavior
\item The current context warrants re-application of previously fulfilled guidelines
\end{enumerate}

When receiving a response from the LLM, the guideline in question can be identified by its ID, and the guideline becomes active or inactive depending on its \texttt{applies\_score}.
When receiving a response from the LLM, the guideline at question can be identified by its ID, and the guideline becomes active or inactive depending on its \texttt{applies\_score}. For guidelines that have been previously applied, we also require the returned value for the key \texttt{guideline\_should\_reapply'} to be \texttt{true}.

\section{Tool Caller Instructions and ARQs}
\label{app:tool-caller}
The tool caller is responsible for determining which tools should be executed based on the current conversation state and active guidelines. Below are the instructions provided to the tool caller in its prompt, along with its expected output format while in ARQ mode:
\begin{lstlisting}
TASK DESCRIPTION
-----------------
Your task is to review the provided tool and, based on your most recent interaction with the customer, decide whether to use it.
For the provided tool, assign a score from 1 to 10 to indicate its usefulness at this time, where a higher score indicates that the tool call should execute.
For any tool with a score of 5 or higher, provide the arguments for activation, following the format in its description.

While doing so, take the following instructions into account:
1. You may suggest tools that don’t directly address the customer’s latest interaction but can advance the conversation to a more useful state based on function definitions.
2. Each tool may be called multiple times with different arguments.
3. Avoid calling a tool with the same arguments more than once, unless clearly justified by the interaction.
4. Ensure each tool call relies only on the immediate context and staged calls, without requiring other tools not yet invoked, to avoid dependencies.
5. Use the "should_run" argument to indicate whether a tool should be executed, meaning it has a high applicability score and either (a) has not been staged with the same arguments, or (b) was staged but needs to be re-executed.
6. If a tool needs to be applied multiple times (each with different arguments), you may include it in the output multiple times.


Produce a valid JSON object according to the following format:
```json
{
    "last_customer_message": "<REPEAT THE LAST USER MESSAGE IN THE INTERACTION>",
    "most_recent_customer_inquiry_or_need": "<customer's inquiry or need>",
    "most_recent_customer_inquiry_or_need_was_already_resolved": <BOOL>,
    "name": "<TOOL NAME>",
    "subtleties_to_be_aware_of": "<NOTE ANY SIGNIFICANT SUBTLETIES TO BE AWARE OF WHEN RUNNING THIS TOOL IN OUR AGENT'S CONTEXT>",
    "tool_calls_for_candidate_tool": [
        {
            "applicability_rationale": "<A FEW WORDS THAT EXPLAIN WHETHER AND HOW THE TOOL NEEDS TO BE CALLED>",
            "applicability_score": <INTEGER FROM 1 TO 10>,
            "argument_evaluations": <EVALUATIONS FOR THE ARGUMENTS. CAN BE DROPPED IF THE TOOL SHOULD NOT EXECUTE>,
            "same_call_is_already_staged": <BOOL>,
            "comparison_with_rejected_tools_including_
            references_to_subtleties": "<A VERY BRIEF OVERVIEW OF HOW THIS CALL FARES AGAINST OTHER TOOLS IN APPLICABILITY>",
            "relevant_subtleties": "<IF SUBTLETIES FOUND, REFER TO THE RELEVANT ONES HERE>",
            "a_rejected_tool_would_have_been_a_better_fit_if_it
            _werent_already_rejected": <BOOL>,
            "potentially_better_rejected_tool_name": "<IF CANDIDATE TOOL IS A WORSE FIT THAN A REJECTED TOOL, THIS IS THE NAME OF THAT REJECTED TOOL>",
            "potentially_better_rejected_tool_rationale": "<IF CANDIDATE TOOL IS A WORSE FIT THAN A REJECTED TOOL, THIS EXPLAINS WHY>",
            "the_better_rejected_tool_should_clearly_be_run_in_
            tandem_with_the_candidate_tool": <BOOL>,
            "should_run": <BOOL>
        }
        ...
    ]
}
```
\end{lstlisting}
These ARQs guide the LLM through an evaluation process that includes:
\begin{enumerate}
\item Identifying the customer's most recent inquiry or need
\item Determining the applicability score of the tool (1-10)
\item Evaluating each required parameter to determine if it is available in context
\item Assessing if the same tool call is already staged
\item Comparing the current tool with other potential tools
\end{enumerate}
When determining parameter values, the tool caller analyzes each parameter's availability and appropriateness using ARQs, checking if:
\begin{itemize}
\item The parameter is provided in the current context
\item The parameter should principally be provided by the customer
\item The parameter was already provided and needs to be provided again
\item It would be problematic to guess the parameter value if not provided
\end{itemize}

A tool is executed if the the LLM's response for it has \texttt{should\_run} set to \texttt{true} in the ARQ response. 

An example of a full tool caller prompt, which includes these instructions in txt format, is available in the supplementary materials.

\section{Message Generator Prompt and ARQs}
\label{app:message-generator}
The Message Generator module is the final component in the processing pipeline, responsible for synthesizing previous module outputs into a coherent response. It follows these instructions, which are provided to it in its prompt:
\begin{lstlisting}
TASK DESCRIPTION:
-----------------
Continue the provided interaction in a natural and human-like manner.
Your task is to produce a response to the latest state of the interaction.
Always abide by the following general principles (note these are not the "guidelines". The guidelines will be provided later):
1. GENERAL BEHAVIOR: Craft responses that feel natural and human-like. Keep them concise and polite, striking a balance between warmth and brevity without becoming overly verbose.
2. AVOID REPEATING YOURSELF: When replying— avoid repeating yourself. Instead, refer the customer to your previous answer, or choose a new approach altogether. If a conversation is looping, point that out to the customer instead of maintaining the loop.
3. DO NOT HALLUCINATE: Do not state factual information that you do not know or are not sure about. If the customer requests information you're unsure about, state that this information is not available to you.
4. ONLY OFFER SERVICES AND INFORMATION PROVIDED IN THIS PROMPT: Do not output information or offer services based on your intrinsic knowledge - you must only represent the business according to the information provided in this prompt.
5. REITERATE INFORMATION FROM PREVIOUS MESSAGES IF NECESSARY: If you previously suggested a solution, a recommendation, or any other information, you may repeat it when relevant. Your earlier response may have been based on information that is no longer available to you, so it’s important to trust that it was informed by the context at the time.
6. MAINTAIN GENERATION SECRECY: Never reveal details about the process you followed to produce your response. Do not explicitly mention the tools, context variables, guidelines, glossary, or any other internal information. Present your replies as though all relevant knowledge is inherent to you, not derived from external instructions.
7. OUTPUT FORMAT: In your generated reply to the customer, use markdown format when applicable.

MESSAGE GENERATION MECHANISM
-----------------
To generate an optimal response that aligns with all guidelines and the current interaction state, follow this structured revision process:

1. INSIGHT GATHERING (Pre-Revision)
   - Before starting revisions, identify up to three key insights from:
     * Explicit or implicit customer requests
     * Relevant principles from this prompt
     * Notable patterns or conclusions from the interaction
   - Each insight should be actionable and directly relevant to crafting the response
   - Only include absolutely necessary insights; fewer is better
   - Document insights' sources for traceability

2. INITIAL RESPONSE
   - Draft an initial response based on:
     * Primary customer needs
     * Applicable guidelines
     * Gathered insights
   - Focus on addressing the core request first

3. REVISION CRITERIA
   The response requires further revision if any of these conditions are met:
   - Facts or services are offered without clear sourcing from this prompt - deonted by all_facts_and_services_sourced_from_prompt being false
   - Guidelines or insights are broken (except when properly prioritized, or when broken due to insufficient data) - denoted by either `instructions_broken_due_to_missing_data` or `instructions_broken_only_due_to_prioritization`
   - The response repeats previous messages - denoted by `is_repeat_message` being true.

4. REVISION DOCUMENTATION
   Document each revision in JSON format including:
   - Complete revised message
   - Facts and sources used
   - Services offered and their sources
   - Guidelines/insights followed and broken
   - Repetition assessment
   - Prioritization decisions and rationales
   - Missing data impacts

5. COMPLETION CRITERIA
   The revision process is complete when either:
   - All guidelines and insights are satisfied, or
   - 5 revisions have been attempted, or
   - Remaining issues are justified by:
     * Explicit prioritization decisions
     * Documented data limitations
     * Customer request conflicts


PRIORITIZING INSTRUCTIONS (GUIDELINES VS. INSIGHTS)
-----------------
Deviating from an instruction (either guideline or insight) is acceptable only when the deviation arises from a deliberate prioritization, based on:
    - Conflicts with a higher-priority guideline (according to their priority scores).
    - Contradictions with a customer request.
    - Lack of sufficient context or data.
    - Conflicts with an insight (see below).
In all other cases, even if you believe that a guideline's condition does not apply, you must follow it.

Guidelines vs. Insights:
Sometimes, a guideline may conflict with an insight you've derived.
For example, if your insight suggests "the customer is vegetarian" but a guideline instructs you to offer non-vegetarian dishes, prioritizing the insight would better align with the business's goals—since offering vegetarian options would clearly benefit the customer.

However, remember that the guidelines reflect the explicit wishes of the business you represent. Deviating from them should only occur if doing so does not put the business at risk.
For instance, if a guideline explicitly prohibits a specific action (e.g., "never do X"), you must not perform that action, even if requested by the customer or supported by an insight.

In cases of conflict, prioritize the business's values and ensure your decisions align with their overarching goals.

\end{lstlisting}

The specific ARQs that the message generator responds to are:

\begin{lstlisting}
```json
{
    "last_message_of_customer": "Hey, can I order a large pepperoni pizza with Sprite?",
    "guidelines": [],
    "context_evaluation": {
        "most_recent_customer_inquiries_or_needs": <str, fill out accordingly>,
        "parts_of_the_context_i_have_here_if_any_with_specific_
        information_on_how_to_address_these_needs": "<fill out accordingly>",
        "topics_for_which_i_have_sufficient_information_and_can
        _therefore_help_with": "<fill out accordingly>",
        "what_i_do_not_have_enough_information_to_help_with
        _with_based_on_the_provided_information_that_i_have": "<fill out accordingly>",
        "was_i_given_specific_information_here_on_how_to
        _address_some_of_these_specific_needs": <BOOL>,
        "should_i_tell_the_customer_i_cannot_help_with_some_
        of_those_needs": <BOOL>
    },
    "insights": "[<Up to 3 original insights to adhere to>]",
    "evaluation_for_each_instruction": [


            {
                "number": 1,
                "instruction": "<Insight #1, if it exists>",
                "evaluation": "<your evaluation of how the insight should be followed>",
                "data_available": "<explanation whether you are provided with the required data to follow this insight now>"
            },
            <Additional entries for all insights>
        
    ],
    "revisions": [
    {
        "revision_number": 1,
        "content": <response chosen after revision 1>,
        "factual_information_provided": [
            {
                "fact": <str, statement of a fact in the suggested response>
                "source": <str, source of the fact - either a specific part of this prompt or something else>
                "is_source_based_in_this_prompt": <BOOL>
            },
            ...
        ],
        "offered_services": [
            {
                "service": <str, statement of a fact in the suggested response>
                "source": <str, source of the fact - either a specific part of this prompt or something else>
                "is_source_based_in_this_prompt": <BOOL>
            },
            ...
        ],
        "instructions_followed": <list of guidelines and insights that were followed>,
        "instructions_broken": <list of guidelines and insights that were broken>,
        "is_repeat_message": <BOOL, indicating whether "content" is a repeat of a previous message by the agent>,
        "followed_all_instructions": <BOOL, whether all guidelines and insights followed>,
        "instructions_broken_due_to_missing_data": <BOOL, optional. Necessary only if instructions_broken_only_due_to_prioritization is true>,
        "missing_data_rationale": <STR, optional. Necessary only if instructions_broken_due_to_missing_data is true>,
        "instructions_broken_only_due_to_prioritization": <BOOL, optional. Necessary only if followed_all_instructions is true>,
        "prioritization_rationale": <STR, optional. Necessary only if instructions_broken_only_due_to_prioritization is true>
        "all_facts_and_services_sourced_from_prompt": <BOOL, if false, you must produce further revisions>,
        "further_revisions_required": <BOOL, true iff either instructions were broken due to invalid reasons, if is_repeat_message is true, or if all_facts_and_services_sourced_from_prompt is false>
    },
    ...
    ]
}
```    
\end{lstlisting}
    
Where text in angled brackets represents our instruction to the LLM, rather than actual text it has to output.

These queries force explicit identification of: 
\begin{enumerate}
    \item Customer needs and available information
    \item guideline applicability with reasoning
    \item fact sourcing with guideline adherence tracking
\end{enumerate}
The message generator includes verifying queries, meaning that its instructed to suggest responses and then evaluates them until a satisfactory response is generated.
This revision process enables self-correction when guidelines are broken or hallucinations detected. The final response of the agent is taken from final revision in the message generator's output.

\backmatter

\bigskip


\end{document}